\documentclass[11pt,a4paper]{article}
\usepackage[nohyperref]{naaclhlt2019}
\usepackage{times}
\usepackage{latexsym}
\usepackage{url}
\usepackage{subcaption}
\usepackage{amssymb}					
\usepackage{amsmath}					
\usepackage{amsthm}						
\usepackage[title]{appendix}

\usepackage[colorinlistoftodos]{todonotes}

\usepackage{verbatim}					
\usepackage[inline]{enumitem}					
\usepackage{tikz-cd}					
\usepackage{braket}						
\usetikzlibrary{cd}						
\usetikzlibrary{shapes,arrows,calc}
\usepackage{adjustbox}					
\usepackage[all,cmtip]{xy}
\usepackage{pgfplots}					
\usepackage[english]{babel}
\usepackage[utf8x]{inputenc}			

\usepackage{comment}
\usepackage[makeroom]{cancel}
\usepackage{imakeidx}					
\usepackage{float}						
\usepackage{booktabs}

\pgfplotsset{width=10cm,compat=1.9}		

\newtheorem{theorem}{Theorem}
\theoremstyle{plain}

\numberwithin{equation}{section}

\theoremstyle{definition}

\newtheorem{Def}[theorem]{Definition}

\newcommand{\R}{\mathbb{R}}				

\newcommand{\bee}{\begin{equation}\begin{aligned}}
\newcommand{\eee}{\end{aligned}\end{equation}}

\newcommand{\fracc}{\frac}				
				
\newcommand{\qwe}{\sqrt}					

\newcommand{\vecc}{\mathbf}

\renewcommand{\rm}{\normalshape}			
\AtBeginDocument{}	
\renewcommand{\Set}[1]{\left\{\,#1\,\right\}}	

\newcommand{\interior}[1]{%
  {\kern0pt#1}^{\hspace{0.5mm}\small\mathrm{o}}%
}

\let\phi\varphi

\aclfinalcopy 
\def\aclpaperid{17} 

\title{Characterizing the impact of geometric properties\\ of word embeddings on task performance}

\author{
    Brendan Whitaker$^1$\thanks{\ \ These authors contributed equally to this work.},\ \ \ \ \ \ 
    Denis Newman-Griffis$^1$\footnotemark[1],\ \ \ \ \ \ 
    Aparajita Haldar$^2$\footnotemark[1],\\
    \textbf{Hakan Ferhatosmanoglu}$^2$\textbf{,}\ \ \ \ \ \ 
    \textbf{Eric Fosler-Lussier}$^1$\\
    $^1$The Ohio State University, Columbus, OH, USA\\
    $^2$University of Warwick, Coventry, UK\\
    \texttt{ \{whitaker.213, newman-griffis.1, fosler-lussier.1\}@osu.edu} \\
    \texttt{ \{aparajita.haldar, h.ferhatosmanoglu\}@warwick.ac.uk}
}

\date{}

\begin{document}
\maketitle

\begin{abstract}
    Analysis of word embedding properties to inform their use in
    downstream NLP tasks has largely been studied by assessing nearest
    neighbors. However, geometric properties of the continuous feature
    space contribute directly to the use of embedding features in
    downstream models, and are largely unexplored.
    We consider four properties of word embedding geometry, namely:
    position relative to the origin, distribution of features in 
    the vector space, global pairwise distances, and local
    pairwise distances. We define a sequence of transformations
    to generate new embeddings that expose subsets of these properties
    to downstream models and evaluate change in task performance to
    understand the contribution of each property to NLP models.
    We transform publicly available pretrained embeddings from
    three popular toolkits (word2vec, GloVe, and FastText)
    and evaluate on a variety of intrinsic tasks, which model
    linguistic information in the vector space, and extrinsic 
    tasks, which use vectors as input to machine learning models. 
    We find that intrinsic evaluations are 
    highly sensitive to absolute position, while
    extrinsic tasks rely primarily on local 
    similarity.
    Our findings suggest that future embedding models and post-processing
    techniques should focus primarily on similarity to nearby points in vector space.
\end{abstract}

\section{Introduction}

Learned vector representations of words, known as word embeddings, have
become ubiquitous throughout natural language processing (NLP) applications.
As a result, analysis of embedding spaces to understand their utility as input features
has emerged as
an important avenue of
inquiry, in order to facilitate proper use of embeddings in downstream NLP tasks.
Many analyses have focused on nearest neighborhoods, as a
viable proxy for semantic information \cite{Rogers2018,Pierrejean2018}.
However, neighborhood-based analysis is limited by the unreliability of
nearest neighborhoods \cite{Wendlandt2018}. Further, it is intended to evaluate
the \textit{semantic content} of embedding spaces, as opposed to characteristics
of the feature space itself.

Geometric analysis offers another recent angle from which to understand the
properties of word embeddings, both in terms of their distribution \cite{Mimno2017}
and correlation with downstream performance \cite{Chandrahas2018}.
Through such geometric investigations, neighborhood-based semantic characterizations are augmented
with information about the continuous feature space of an embedding.
Geometric features offer a more direct connection to the assumptions
made by neural models about continuity in input spaces \cite{Szegedy2014},
as well as the use of recent contextualized representation methods
using continuous language models \cite{Peters2018a,Devlin2018}.

\begin{figure*}[t]
    \centering
    \includegraphics[width=0.85\textwidth]{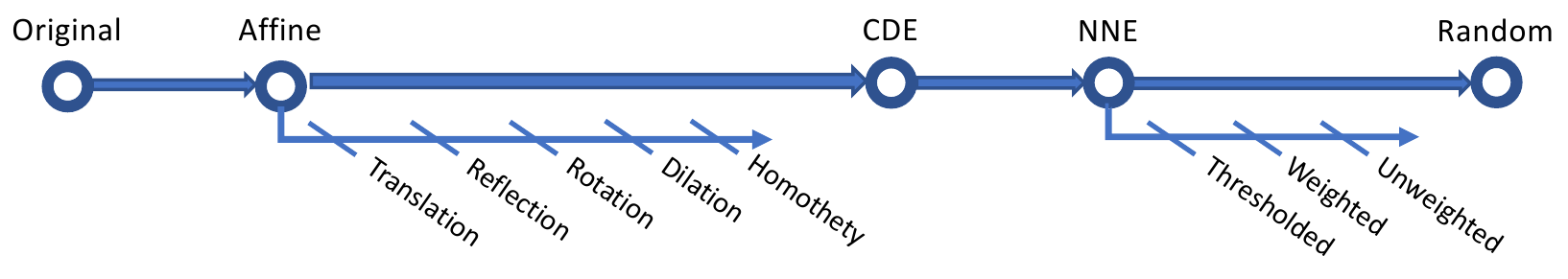}
    \caption{Sequence of transformations applied to word embeddings, including
             transformation variants. Note that each transformation is applied
             independently to source word embeddings. Transformations are
             presented in order of decreasing geometric information retained
             about the original vectors.}
    \label{fig:hierarchy}
\end{figure*}

In this work, we aim to bridge the gap between
neighborhood-based semantic analysis and
geometric performance analysis. We consider four components of
the geometry of word embeddings, and transform pretrained embeddings to expose
only subsets of these components to downstream models.
We transform three popular sets of embeddings,
trained using word2vec \cite{mikolov2013efficient},\footnote{
    3M 300-d GoogleNews vectors from {\small \url{https://code.google.com/archive/p/word2vec/}}
}
GloVe \cite{Pennington2014},\footnote{
    2M 300-d 840B Common Crawl vectors from {\small \url{https://nlp.stanford.edu/projects/glove/}}
}
and FastText \cite{Bojanowski2017},\footnote{
    1M 300-d WikiNews vectors with subword information from {\small \url{https://fasttext.cc/docs/en/english-vectors}}
}
and use the resulting embeddings in a battery of standard
evaluations to measure changes in task performance.

We find that intrinsic evaluations, which model linguistic information directly
in the vector space, are highly sensitive to absolute position in pretrained
embeddings; while extrinsic tasks, in which word embeddings are passed as input
features to a trained model, are more robust and rely primarily on information
about local similarity between word vectors. Our findings, including evidence
that global organization of word vectors is often a major source of noise,
suggest that further development of embedding learning and tuning methods should
focus explicitly on local similarity, and help to explain the success of several
recent methods.

\section{Related Work}

Word embedding models and outputs have been analyzed from
several angles.
In terms of performance,
evaluating the ``quality'' of word
embedding models has long been a thorny problem. While intrinsic
evaluations such as word similarity and analogy completion are
intuitive and easy to compute, they are limited by both confounding
geometric factors \cite{Linzen2016} and task-specific factors
\cite{Faruqui2016, Rogers2017}. \citet{Chiu2016a} show that these
tasks, while correlated with some semantic content, do not always 
predict downstream performance. Thus, it is necessary to use a
more comprehensive set of intrinsic and extrinsic evaluations for
embeddings.

Nearest neighbors in sets of embeddings are commonly used as a proxy for qualitative
semantic information. However, their instability across embedding
samples \cite{Wendlandt2018} is a limiting factor, and they do not
necessarily correlate with linguistic analyses \cite{Hellrich2016}.
Modeling neighborhoods as a graph structure offers an alternative
analysis method \cite{Gyllensten2015}, as does 2-D or 3-D visualization 
\cite{Heimerl2018}. However, both of these methods provide qualitative
insights only. By systematically analyzing geometric information with
a wide variety of evaluations, we provide a quantitative counterpart
to these understandings of embedding spaces.

\section{Methods}
\label{sec:methods}

In order to investigate how different geometric properties of word embeddings
contribute to model performance on intrinsic and extrinsic evaluations, 
we consider the following attributes of word embedding geometry:
\vspace{-0.2cm}
\begin{itemize}
    \setlength\itemsep{-0.3em}
    \item position relative to the origin;
    \item distribution of feature values in $\mathbb{R}^d$;
    \item global pairwise distances, i.e.\ distances between any pair of vectors;
    \item local pairwise distances, i.e.\ distances between nearby pairs of vectors.
    \vspace{-0.2cm}
\end{itemize}

Using each of our sets of pretrained word embeddings, we apply a variety
of transformations to induce new embeddings that only expose subsets of these
attributes to downstream models. These are: affine transformation, which
obfuscates the original position of the origin; cosine distance encoding,
which obfuscates the original distribution of feature values in $\R^d$;
nearest neighbor encoding, which obfuscates global pairwise distances;
and random encoding. This sequence is illustrated in Figure~\ref{fig:hierarchy},
and the individual transformations are discussed in the following subsections.

General notation for defining our transformations is as follows.
Let $W$ be our vocabulary of words 
taken from some source corpus. We associate
with each word $w \in W$ a vector $\vecc{v} \in \R^d$ resulting from training
via one of our embedding generation algorithms, where $d$ is an arbitrary
dimensionality for the embedding space. We define
$V$ to be the set of all pretrained word vectors $\vecc{v}$ for a given
corpus, embedding algorithm, and parameters. The matrix 
of embeddings $M_V$ associated with this set then has shape 
$|V| \times d$.
For simplicity, we restrict our analysis to transformed embeddings
of the same dimensionality $d$ as the original vectors.

\subsection{Affine transformations}
\label{affinefuncs}

Affine transformations have been previously utilized for post-processing of word embeddings.
For example, \citet{Artetxe2016} learn a matrix transform to align
multilingual embedding spaces, and \citet{Faruqui2015b} use a linear sparsification
to better capture lexical semantics. In addition, the simplicity of
affine functions in machine learning contexts \cite{Hofmann2008} makes them a
good starting point for our analysis.

Given a set of embeddings in $\mathbb{R}^d$, referred to as an 
\textbf{embedding space}, affine transformations 
\begin{equation*}
    f_{\textrm{affine}} : \mathbb{R}^{d} \rightarrow \mathbb{R}^{d} 
\end{equation*}
change positions of points relative to the origin.

While prior work has typically focused on linear transformations, which fix the
origin, we consider the broader class of affine transformations, which do not.
Thus, affine transformations
such as translation cannot in general be represented as a square matrix for
finite-dimensional spaces.

We use the following affine transformations:
\vspace{-0.3em}
\begin{itemize}
\setlength\itemsep{-0.3em}
    \item translations;
    \item reflections over a hyperplane; 
    \item rotations about a subspace; 
    \item homotheties.
\end{itemize}
We give brief definitions of each transformation.
\begin{Def}
    A \textbf{translation} is a function $T_{\vecc{x}}:\R^d \to \R^d$
    given by 
    \vspace{-0.1cm}
    \bee
    T_{\vecc{x}}(\vecc{v}) = \vecc{v} + \vecc{x}
    \eee
    \vspace{-0.1cm}
    where $\vecc{x} \in \R^d$. 
\end{Def}

\begin{Def}
    For every $\vecc{a} \in \R^d$, we call the map
    $\mathrm{Refl}_{\vecc{a}}:\R^d \to \R^d$ given by
    \vspace{-0.1cm}
    \bee
        \mathrm{Refl}_{\vecc{a}}(\vecc{v}) = \vecc{v} - 2\frac{\vecc{v}\cdot \vecc{a}}{\vecc{a} \cdot \vecc{a}}\vecc{a}
    \eee 
    \vspace{-0.1cm}
    the \textbf{reflection} over the hyperplane through the origin orthogonal
    to $\vecc{a}$. 
\end{Def}

\begin{Def}
    A \textbf{rotation} through the span of vectors 
    $\vecc{u},\vecc{x}$ by angle $\theta$ is a map 
    $\mathrm{Rot}_{\vecc{u},\vecc{x}}: \R^d \to \R^d$ given by
    \vspace{-0.1cm}
    \bee
    	\mathrm{Rot}_{\vecc{u},\vecc{x}}(\vecc{v})
    	 = A\vecc{v}
    \vspace{-0.1cm}
    \eee
    where 
    \vspace{-0.1cm}
    \bee
    	A &= I + \sin\theta(\vecc{x}\vecc{u}^T - \vecc{u}\vecc{x}^T)
    	\\
    	&+ (\cos\theta - 1)(\vecc{u}\vecc{u}^T + 
    	\vecc{x}\vecc{x}^T)
    \vspace{-0.1cm}
    \eee
    and $I \in \mathrm{Mat}_{d,d}(\R)$ is the identity matrix. 
    
\end{Def}

\begin{Def}
    For every $\vecc{a} \in \R^d$ and $\lambda \in \R \setminus \Set{0}$, 
    we call the map  $H_{\vecc{a}, \lambda}:\R^d \to \R^d$ given by
    \bee
        H_{\vecc{a}, \lambda}(\vecc{v}) = \vecc{a} + \lambda (\vecc{v} - \vecc{a})
    \eee
    a \textbf{homothety} of center $\vecc{a}$ and ratio $\lambda$. A homothety centered at the origin is called a \textbf{dilation}.
\end{Def}

Parameters used in our analysis for each of these transformations are
provided in Appendix~\ref{app:parameters}.

\subsection{Cosine distance encoding (CDE)}
\label{aesect}

Our cosine distance encoding transformation
\begin{equation*}
    f_{\textrm{CDE}} : \mathbb{R}^{d} \rightarrow \mathbb{R}^{|V|}
\end{equation*}
obfuscates the distribution of features in $\mathbb{R}^d$
by representing a set of word vectors as a pairwise distance
matrix. Such a transformation might be used to avoid the
non-interpretability of embedding features \cite{Fyshe2015} and
compare embeddings based on relative organization alone.

\begin{Def}
Let $\vecc{a},\vecc{b} \in \R^d$. Then their \textbf{cosine distance}
$d_{\cos}:\R^d \times \R^d \to [0,2]$ is given by
\begin{equation}
	d_{\cos}(\vecc{a},\vecc{b}) 
	= 1 - \fracc{\vecc{a}\cdot \vecc{b}}{||\vecc{a}||||\vecc{b}||} 
    \label{eq:cosine-distance}
\end{equation}
where the second term is the \textbf{cosine similarity}.
\end{Def}

As all three sets of embeddings evaluated in this study have vocabulary
size on the order of $10^6$, use of the full distance matrix is impractical. 
We use a subset consisting of the distance from each point to
the embeddings of the 10K most frequent words from each embedding set,
yielding
\begin{equation*}
    f_{\textrm{CDE}} : \R^d \rightarrow \R^{10^4}
\end{equation*}
This is not dissimilar to the global frequency-based negative sampling
approach of word2vec \cite{mikolov2013efficient}.
We then use an autoencoder to map this
back to
$\mathbb{R}^d$ for comparability.
\begin{Def}
Let $\vecc{v}\in \R^{|V|}, \vecc{W_1},\vecc{W_2}\in\R^{|V|\times d}$.
Then an \textbf{autoencoder} over $\R^{|V|}$ is defined as
\begin{align}
    \vecc{h} &= \phi(\vecc{v}\vecc{W_1})\\
    \hat{\vecc{v}} &= \phi(\vecc{W_2}^T\vecc{h})
\end{align}
Vector $\vecc{h}\in\R^d$ is then used as the compressed representation of $\vecc{v}$.
\end{Def}

In our experiments, we use ReLU as our activation function $\phi$, and
train the autoencoder for 50 epochs to minimize $L^2$ distance between $\vecc{v}$
and $\hat{\vecc{v}}$.

We recognize that low-rank compression using an autoencoder is likely to be noisy,
thus potentially inducing additional loss in evaluations.
However, precedent for capturing geometric structure with autoencoders
\cite{Li2017grass} suggests that this is a viable model for our analysis.

\subsection{Nearest neighbor encoding (NNE)}

Our nearest neighbor encoding transformation
\begin{equation*}
    f_{\textrm{NNE}} : \R^d \rightarrow \R^{|V|}
\end{equation*}
discards the majority of the global
pairwise distance information modeled in CDE,
and retains only information about nearest neighborhoods.
The output of $f_{\textrm{NNE}}(\vecc{v})$ is a
sparse vector.

This transformation relates to the common use of nearest neighborhoods
as a proxy for semantic information
\cite{Wendlandt2018,Pierrejean2018}. We take the previously proposed
approach of combining the output of $f_{\textrm{NNE}}(\vecc{v})$ for
each $\vecc{v} \in V$ to form a sparse adjacency matrix, which describes a
directed nearest neighbor graph \cite{Gyllensten2015,Newman-Griffis2017second},
using three versions of $f_{\textrm{NNE}}$ defined below.

\textbf{Thresholded} The set of non-zero indices in $f_{\textrm{NNE}}(\vecc{v})$
correspond to word vectors $\tilde{\vecc{v}}$ such that the cosine similarity
of $\vecc{v}$ and $\tilde{\vecc{v}}$ is greater than or equal to an arbitrary
threshold $t$. In order to ensure that every word has non-zero out degree in
the graph, we also include the $k$ nearest neighbors by cosine similarity for
every word vector. Non-zero values in $f_{\textrm{NNE}}(\vecc{v})$ are set to
the cosine similarity of $\vecc{v}$ and the relevant neighbor vector.

\textbf{Weighted} The set of non-zero indices in $f_{\textrm{NNE}}(\vecc{v})$
corresponds to only the set of $k$ nearest neighbors to $\vecc{v}$ by cosine
similarity. Cosine similarity values are used for edge weights.

\textbf{Unweighted} As in the previous case, only $k$ nearest neighbors are included in the adjacency matrix. All edges are
weighted equally, regardless of cosine similarity.

We report results using $k=5$ and $t=0.05$; other settings are discussed in Appendix~\ref{app:nne}.

Finally, much like the CDE method, we use a second mapping function
\begin{equation*}
    \psi : \R^{|V|} \rightarrow \R^d
\end{equation*}
to transform the nearest neighbor graph back to $d$-dimensional vectors for
evaluation. Following \citet{Newman-Griffis2017second}, we use node2vec
\cite{Grover2016} with default parameters to learn this mapping. Like
the autoencoder, this is a noisy map, but the intent of node2vec to capture
patterns in local graph structure makes it a good fit for our analysis.

\begin{figure*}[t]
    \centering
    \begin{subfigure}[b]{\textwidth}
    \includegraphics[width=\textwidth]{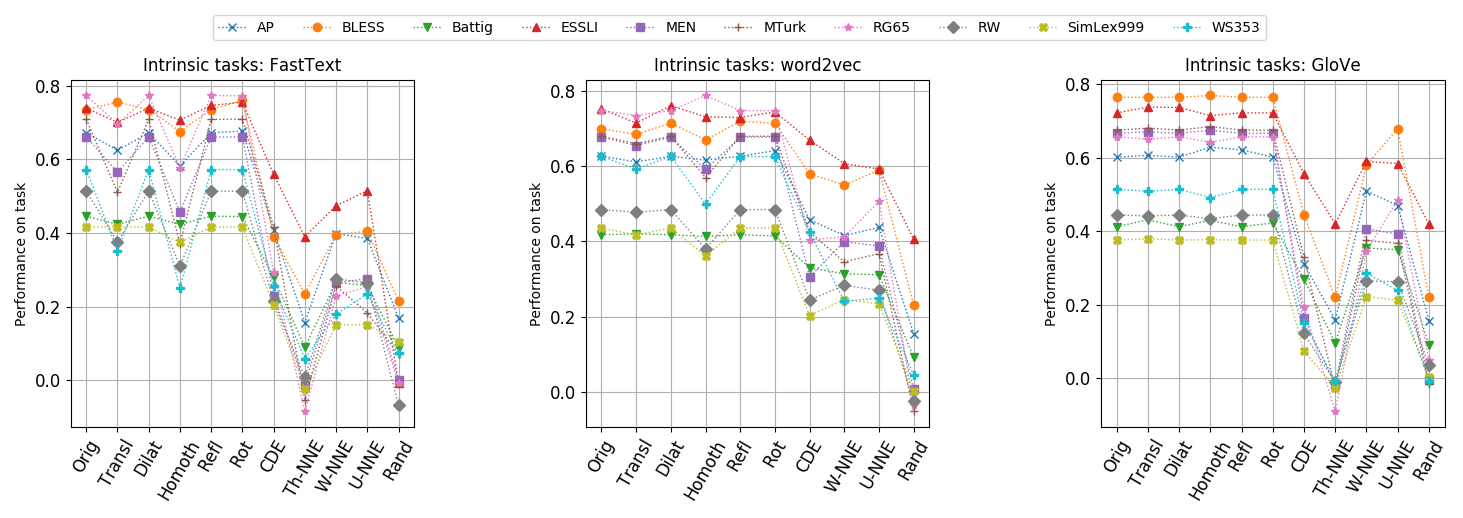}
    \caption{Results of intrinsic evaluations}
    \end{subfigure}
    \begin{subfigure}[b]{\textwidth}
    \includegraphics[width=\textwidth]{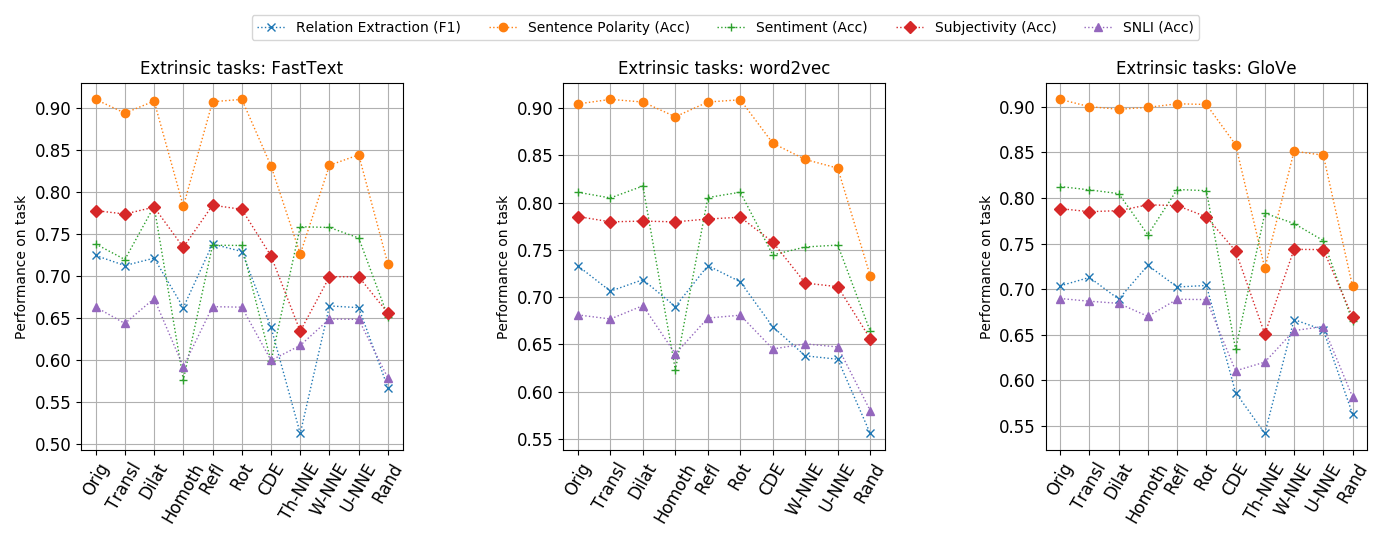}
    \caption{Results of extrinsic evaluations}
    \end{subfigure}
    \caption{Performance metrics on intrinsic and extrinsic tasks,
             comparing across different transformations applied to
             each set of word embeddings. Dotted lines are for visual aid in tracking performance on individual tasks, and do not indicate continuous transformations. Transformations are presented in order of decreasing geometric information about the original vectors, and are applied independent of one another to the original source embedding.}
    \label{fig:results}
\end{figure*}

\subsection{Random encoding}

Finally, as a baseline, we use a random encoding
\begin{equation*}
    f_{\textrm{Rand}} : \R^d \rightarrow \R^d
\end{equation*}
that discards original vectors entirely. 

While intrinsic evaluations rely only on input embeddings, and
thus lose all source information in this case, extrinsic tasks
learn a model to transform input features, making even randomly-initialized
vectors a common baseline \cite{lample2016neural,Kim2014}. For fair
comparison, we generate one set of random baselines for each embedding
set and re-use these across all tasks.

\subsection{Other transformations}

Many other transformations of a word embedding space could be included
in our analysis, such as arbitrary vector-valued polynomial functions, rational vector-valued functions, 
or common decomposition methods such as principal components analysis (PCA)
or singular value decomposition (SVD). Additionally, though they cannot
be effectively applied to the unordered set of word vectors in a raw embedding space,
transformations for sequential data such as discrete Fourier transforms or
discrete wavelet transforms could be used for word sequences in specific
text corpora.

For this study, we limit our scope to the transformations
listed above. These transformations align with prior work on analyzing
and post-processing embeddings for specific tasks, and are highly interpretable
with respect to the original embedding space. However, other complex transformations
represent an intriguing area of future work.

\section{Evaluation}

In order to measure the contributions of each geometric aspect described in
Section~\ref{sec:methods} to the utility of word embeddings as input features,
we evaluate embeddings transformed using our sequence of operations on a battery of
standard intrinsic evaluations, which model linguistic information directly in the
vector space; and extrinsic evaluations, which use the embeddings as input to learned
models for downstream applications
Our intrinsic evaluations include:
\vspace{-0.3em}
\begin{itemize}
\setlength\itemsep{-0.3em}
    \item Word similarity and relatedness, using cosine similarity:
          WordSim-353 \cite{WordSim353}, SimLex-999 \cite{SimLex-999},
          RareWords \cite{Luong2013}, RG65 \cite{Rubenstein1965},
          MEN \cite{Bruni2014}, and MTURK \cite{Radinsky2011}.\footnote{
              \url{https://github.com/kudkudak/word-embeddings-benchmarks}
              using single-word datasets only.
              For brevity, we omit the Sim/Rel splits of WordSim-353
              \cite{Agirre2009}, which showed the same trends as the full dataset.
          }
    \item Word categorization, using an oracle combination of agglomerative
          and $k$-means clustering: AP \cite{Almuhareb2005}, BLESS \cite{BLESS},
          Battig \cite{Battig1969}, and the ESSLLI 2008 shared task
          (\citet{ESSLLI2008}, performance averaged across
          nouns, verbs, and concrete nouns).\footnotemark[5]
\end{itemize}

Given the well-documented issues with using vector arithmetic-based
analogy completion as an intrinsic evaluation \cite{Linzen2016,Rogers2017,Newman-Griffis2017},
we do not include it in our analysis.

We follow \citet{Rogers2018} in evaluating on a set of five extrinsic tasks:\footnote{
    \url{https://github.com/drgriffis/Extrinsic-Evaluation-tasks}
}

\vspace{-0.5em}
\begin{itemize}
\setlength\itemsep{-0.3em}
    \item Relation classification: SemEval-2010 Task 8 \cite{Hendrickx2010},
          using a CNN with word and distance embeddings \cite{Zeng2014}.
    \item Sentence-level sentiment polarity classification: MR movie reviews
          \cite{Pang2005}, with a simplified CNN model from \cite{Kim2014}.
    \item Sentiment classification: IMDB movie reviews \cite{Maas2011}, with
          a single 100-d LSTM.
    \item Subjectivity/objectivity classification: Rotten Tomato snippets
          \cite{Pang2004}, using a logistic regression over summed word
          embeddings \cite{Li2017}.
    \item Natural language inference: SNLI \cite{SNLI}, using separate
          LSTMs for premise and hypothesis, combined with a feed-forward
          classifier.
\end{itemize}

\section{Analysis and Discussion}

Figure \ref{fig:results} presents the results of each intrinsic and extrinsic
evaluation on the transformed versions of our three sets of word embeddings.\footnote{
    Due to their large vocabulary size, we were unable to run Thresholded-NNE
    experiments with word2vec embeddings.
}
The largest drops in performance across all three sets for intrinsic tasks
occur when explicit embedding features are removed with the CDE transformation.
While some cases of NNE-transformed embeddings recover a measure of this performance,
they remain far under affine-transformed embeddings. Extrinsic tasks are similarly
affected by the CDE transformation; however, NNE-transformed embeddings recover
the majority of performance.

Comparing within the set of affine transformations, the innocuous effect of
rotations, dilations,
and reflections on both intrinsic and extrinsic tasks suggests that
the models used are robust to simple linear transformations. Extrinsic
evaluations are also relatively insensitive to translations, which can be modeled
with bias terms, though the lack of learned models and reliance
on cosine similarity for the intrinsic tasks makes them more sensitive to
shifts relative to the origin.
Interestingly, homothety, which effectively combines
a translation and a dilation, leads to a noticeable drop in performance across
all tasks. Intuitively, this result makes sense: by both shifting points relative
to the origin and changing their distribution in the space, angular similarity
values used for intrinsic tasks can be changed significantly, and the zero mean
feature distribution preferred by neural models \cite{Clevert2016} becomes harder
to achieve.
This suggests that methods for tuning embeddings should
attempt to preserve the origin whenever possible.

The large drops in performance observed when using the CDE transformation is likely
to relate to the instability of nearest neighborhoods and the importance of locality
in embedding learning \cite{Wendlandt2018}, although the effects of the autoencoder component also bear further investigation.
By effectively increasing the size of the
neighborhood considered, CDE adds additional sources of semantic noise. The similar
drops from thresholded-NNE transformations, by the same token, is likely related to
observations of the relationship
between the frequency ranks of a word and its nearest neighbors \cite{Faruqui2016}.
With thresholded-NNE, we find that the words with highest out degree in the
nearest neighbor graph are rare words (e.g., ``Chanterelle'' and ``Courtier'' in
FastText, ``Tiegel'' and ``demangler'' in GloVe), which link to other rare words.
Thus, node2vec's random walk method is more likely to traverse these
dense subgraphs of rare words, adding noise to the output embeddings.

Finally, we note that \citet{Melamud2016} showed significant
variability in downstream task performance when using different
embedding dimensionalities. While we fixed vector dimensionality
for the purposes of this study, varying $d$ in future work 
represents a valuable follow-up.

Our findings suggest that methods for training and tuning embeddings, especially
for downstream tasks, should explicitly focus on local geometric structure in the vector
space.
One concrete example of this comes from \citet{Chen2018}, who demonstrate empirical
gains when changing the negative sampling approach of word2vec to choose negative
samples that are currently near to the target word in vector space, instead of the
original frequency-based sampling (which ignores geometric structure). Similarly,
successful methods for tuning word embeddings for specific tasks have often focused
on enforcing a specific neighborhood structure \cite{Faruqui2015b}. We demonstrate
that by doing so, they align qualitative semantic judgments with the primary geometric
information that downstream models learn from.

\section{Conclusion}

Analysis of word embeddings has largely focused on qualitative
characteristics such as nearest neighborhoods or relative distribution.
In this work, we take a quantitative approach analyzing geometric
attributes of embeddings in $\R^d$, in order to understand the impact
of geometric properties on downstream task performance. We characterized
word embedding geometry in terms of absolute position, vector features,
global pairwise distances, and local pairwise distances, and generated
new embedding matrices by removing these attributes from pretrained
embeddings. By evaluating the performance of these transformed embeddings
on a variety of intrinsic and extrinsic tasks, we find that while intrinsic
evaluations are sensitive to absolute position, downstream models rely
primarily on information about local similarity.

As embeddings are used for increasingly specialized applications, and
as recent contextualized embedding methods such as ELMo \cite{Peters2018a}
and BERT \cite{Devlin2018} allow
for dynamic generation of embeddings from specific contexts, our findings
suggest that work on tuning and improving these embeddings should focus
explicitly on local geometric structure in sampling and evaluation
methods.
The source code for our transformations and complete tables of
our results are available online at \url{https://github.com/OSU-slatelab/geometric-embedding-properties}.

\section*{Acknowledgments}
We gratefully acknowledge the use of Ohio Supercomputer Center \cite{OSC}
resources for this work, and thank our anonymous reviewers for their insightful
comments. Denis is supported via a Pre-Doctoral Fellowship from the National
Institutes of Health, Clinical Center. Aparajita is supported via a Feuer International Scholarship in Artificial Intelligence.

\bibliographystyle{acl_natbib}
\bibliography{naaclhlt2019}

\begin{appendices}

\section{Parameters}
\label{app:parameters}

We give the following library of vectors in $\R^d$ 
used as parameter values:
\bee
    \vecc{v}_{\mathrm{diag}} &= 
        \begin{bmatrix}
            \frac{1}{\qwe{d}} \\
            \vdots \\
            \frac{1}{\qwe{d}}
        \end{bmatrix} ;
        \\
    \vecc{v}_{\mathrm{diagNeg}} &=
        \begin{bmatrix}
           - \frac{1}{\qwe{d}} \\
            \frac{1}{\qwe{d}} \\
            \vdots \\
            \frac{1}{\qwe{d}}
        \end{bmatrix}. 
        \\
\eee

\begin{table}[H]
\begin{adjustbox}{center}
    \begin{tabular}{@{}lll@{}}
    \toprule
	\textbf{Transform} & \textbf{Parameter} & \textbf{Value}\\
    \midrule
    Translation & Direction:& $\vecc{0}$ \\
    & Magnitude:& $1$\vspace{2mm}\\
    Dilation & Magnitude:& $2$\vspace{2mm}\\
    Homothety & Center:& $\vecc{v}_{\mathrm{diag}}$\\ 
    & Magnitude:& $0.25$\vspace{2mm}\\
    Reflection & Hyperplane Vector:& $\vecc{v}_{\mathrm{diag}}$\vspace{2mm}\\
    2-D Rotation & Basis Vector 1:& $\vecc{v}_{\mathrm{diag}}$ \\
    & Basis Vector 2:& $\vecc{v}_{\mathrm{diagNeg}}$ \\
    & Angle:& $\pi/4$\\
    \bottomrule
    \end{tabular}%
  \label{tab3}%
  \end{adjustbox}
  \caption{Transform parameters. }
\end{table}

\section{NNE settings}
\label{app:nne}

We experimented with $k\in\{5,10,15\}$ for our weighted and unweighted
NNE transformations. For thresholded NNE, in order to best evaluate the
impact of thresholding over uniform $k$, we used the minimum $k=5$ and
experimented with $t\in\{0.01,0.05,0.075\}$; higher values of $t$ increased
graph size sufficiently to be impractical. We report using $k=5$ for weighted
and unweighted settings in our main results for fairer comparison with the
thresholded setting.

The effect of thresholding on nearest neighbor graphs was a strongly
right-tailed increase in out degree for a small portion of nodes.
Our reported value of $t=0.05$ increased the out degree of 20,229 nodes
for FastText (out of 1M total nodes), with the maximum increase being
819 (``Chanterelle''), and 1,354 nodes increasing out degree by only 1.
For GloVe, 7,533 nodes increased in out degree (out of 2M total), with
maximum increase 240 (``Tiegel''), and 372 nodes increasing out degree
by only 1.

Table~\ref{tbl:nne-intrinsic} compares averaged performance values across
all intrinsic tasks for these settings, and Table~\ref{tbl:nne-extrinsic}
compares average extrinsic task performance.

\begin{table}[h]
    \centering
    {\small
    \begin{tabular}{r|ccc}
        NNE params&FastText&word2vec&GloVe\\
        \hline
        \multicolumn{4}{l}{\textit{Thresholded}}\\
        \hline
        $k=5,t=0.01$&0.160&--&0.106\\
        $k=5,t=0.05$&0.129&--&0.130\\
        $k=5,t=0.075$&0.150&--&0.132\\
        \hline
        \multicolumn{4}{l}{\textit{Weighted}}\\
        \hline
        $k=5$&0.320&0.419&0.426\\
        $k=10$&0.342&0.363&0.460\\
        $k=15$&0.346&0.376&0.448\\
        \hline
        \multicolumn{4}{l}{\textit{Unweighted}}\\
        \hline
        $k=5$&0.330&0.428&0.435\\
        $k=10$&0.351&0.396&0.463\\
        $k=15$&0.341&0.365&0.432\\
    \end{tabular}
    }
    \caption{Mean performance on intrinsic tasks under different NNE
             settings.}
    \label{tbl:nne-intrinsic}
\end{table}

\begin{table}[h]
    \centering
    {\small
    \begin{tabular}{r|ccc}
        NNE params&FastText&word2vec&GloVe\\
        \hline
        \multicolumn{4}{l}{\textit{Thresholded}}\\
        \hline
        $k=5,t=0.01$&0.642&--&0.666\\
        $k=5,t=0.05$&0.650&--&0.664\\
        $k=5,t=0.075$&0.649&--&0.663\\
        \hline
        \multicolumn{4}{l}{\textit{Weighted}}\\
        \hline
        $k=5$&0.721&0.720&0.738\\
        $k=10$&0.728&0.713&0.740\\
        $k=15$&0.725&0.713&0.739\\
        \hline
        \multicolumn{4}{l}{\textit{Unweighted}}\\
        \hline
        $k=5$&0.720&0.717&0.732\\
        $k=10$&0.724&0.712&0.738\\
        $k=15$&0.729&0.708&0.725\\
    \end{tabular}
    }
    \caption{Mean performance on extrinsic tasks under different NNE
             settings.}
    \label{tbl:nne-extrinsic}
\end{table}

\end{appendices}

\end{document}